\documentstyle[12pt,a4]{article}

\begin{document}

\baselineskip 18pt

\begin{center}
{\bf TRAINING REINFORCEMENT NEUROCONTROLLERS \\
USING THE POLYTOPE ALGORITHM}
\end{center}

\begin{center}
Aristidis Likas and Isaac Lagaris \\
Department of Computer Science \\
University of Ioannina \\
P.O. Box. 1186 - GR 45110 Ioannina, Greece \\
\end{center}

\vspace*{0.5cm}
\begin{center}
Correspondence: A. Likas \\
Department of Computer Science \\
University of Ioannina \\
P.O. Box. 1186 - GR 45110 Ioannina, Greece \\
tel: +30-651-97310\\
fax: +30-651-48131\\
e-mail: arly@cs.uoi.gr
\end{center}

\vspace*{0.5cm}
\begin{abstract}
A new training algorithm is presented for delayed reinforcement
learning problems that does not assume the existence of
a critic model and
employs the polytope optimization algorithm to adjust the weights of
the action network so that a simple direct measure of the training
performance is maximized. Experimental results from the application of
the method to the pole balancing problem indicate improved training
performance compared with critic-based and genetic reinforcement approaches.
\end{abstract}


\pagebreak

\begin{center}
{\bf TRAINING REINFORCEMENT NEUROCONTROLLERS \\
USING THE POLYTOPE ALGORITHM}
\end{center}

\begin{abstract}
A new training algorithm is presented for delayed reinforcement
learning problems that does not assume the existence of
a critic model and
employs the polytope optimization algorithm to adjust the weights of
the action network so that a simple direct measure of the training
performance is maximized. Experimental results from the application of
the method to the pole balancing problem indicate improved training
performance compared with critic-based and genetic reinforcement approaches.
\end{abstract}


\section{Introduction}

In the framework of delayed  reinforcement learning, a system receives
input from its environment, decides for a proper sequence of actions,
executes them, and thereafter receives a reinforcement signal, namely
a grade for the made decision.
A system at any instant is described by its, so called, state variables.
The objective of a broad class of reinforcement problems, 
is to learn 
how to control a system in such a way, so that its state variables remain
at all times within prescribed ranges. However, if at any instant, the system
violates this requirement, it is penalized by receiving a "bad grade" signal,
and hence its policy in making further decisions is influenced accordingly.

There are many examples of this kind of problems, like 
the pole balancing problem, teaching an autonomous robot 
to avoid obstacles, the ball and beam problem \cite{Lin96} etc.

In general we can distinguish two kinds of approaches 
that have been developed for
delayed reinforcement problems \cite{Kae96}: 
the {\em critic-based} approaches and
the {\em direct} approaches. There is also the
{\em Q-learning} approach \cite{Wat92} which exhibits
many similarities with the critic-based ones.
The most well-studied critic-based approach is the 
Adaptive Heuristic Critic (AHC) \cite{And89,And87,Bar83} method which  
assumes
two separate models: an {\em action model} that receives the current
system state and selects the action to be taken and the {\em
evaluation model} which provides as output a prediction $e(x)$ of the
evaluation of the current state $x$. The evaluation model is usually a 
feedforward neural network trained using the method of temporal
differences, i.e. it tries to minimize the error $\delta = e(x) -
(r + \gamma e(y))$ where $y$ is the new state, $r$ the received 
reinforcement 
and $\gamma$ a discount factor \cite{Bar83,And87,And89}.
The action model is also a feedforward network
that provides as output a vector of probabilities upon which the
action selection is based. Both networks are trained on-line 
through backpropagation    
using the same error value $\delta$ described previously.

The {\em direct approach} to delayed reinforcement learning problems
considers reinforcement learning as a general optimization problem
with an objective function having a straightforward 
formulation but which is difficult to optimize \cite{Kae96}. 
In such a case only the action model is necessary to provide the 
action policy 
and optimization techniques must be employed to adjust the
parameters of the action model so that a stochastic 
integer-valued function is maximized. This function is actually 
proportional to the number of 
successful decisions (i.e. actions that do not lead to the receipt of
penalty signal).  
A previous direct approach to delayed reinforcement problems employs  
real-valued genetic algorithms to perform the optimization task \cite{Whi93}.  
In the present study we propose another optimization strategy that is 
based on the polytope method with random restarts. 
Details concerning such an approach are presented
in the next section, while section 3 provides experimental results from
the application of the proposed method to the pole balancing problem and
compares its  performance against that of the AHC method and of the 
evolutionary approach. 

\section{The Proposed Training Algorithm}

As already mentioned, 
the proposed method belongs to the category
of direct approaches to delayed reinforcement problems.
Therefore, only an action model is considered that in our case has
the architecture of a 
multilayer perceptron with input units accepting the system state
at each time instant,
and sigmoid output units providing
output values $p_i$ in the range $(0, 1)$. The decision for the action
to be taken from the values of $p_i$ can be made either stochastically 
or deterministically. For example in the case of one output unit
the value $p$ may represent the probability that the final output will be
one or zero, or the final output may be obtained deterministically using
the rule: if $p > 0.5$ the final output will be one, 
otherwise it will be zero.   
Learning proceeds in {\em cycles}, with each cycle starting with the system 
placed  at a random initial position and ending with a failure signal.
Since our objective is to train the network so that the system ideally never
receives a failure signal,
the number of time steps of the cycle (ie. its length),
constitutes the performance measure to be optimized. 
Consequently, the training problem
can be considered as a function optimization problem with 
the adjustable parameters being the weights and biases of the action network
and with the function value being the length of a cycle obtained
using the current weight values.
In practice, when the length of a cycle exceeds a preset maximum number 
of steps, we consider that the controller has been adequately trained. 
This is used as a criterion for terminating the training process.
The training also terminates if the number of unsuccessful cycles (i.e.
function evaluations without reaching maximum value) 
exceed a preset upper bound. 

Obviously, 
the function to be optimized is integer-valued, 
thus it is not possible to define
derivatives.
Therefore, traditional gradient-based
optimization techniques cannot be employed. 
Moreover, the function posesses an amount of random noise
since the initial state specification as well as the action selection
at the early steps are performed at random.
On one hand
the incorporation of this random noise may disrupt the optimization process.
For example, if the evaluation of the same network is radically different 
at different times, then the learning process will be misled. 
On the other hand, the global search certainly benefits from it
and hence the noise should be kept, however under control. \\
It is clear that the direct approach has certain advantages which we 
summarize in the following list.
\begin{itemize}
\item Instead of using an on-line update strategy for the
action network, we perform updates only at the end of each cycle.
Therefore, the policy of the action network is not affected
in the midst of a cycle (during which the network actually performs well).
The continuous on-line adjustment
of the weights of  the action network may lead due to overfitting,
to the corruption of correct policies that the system has acquired 
so far \cite{Kon94}.
\item Several sofisticated, derivative-free, multidimensional 
      optimization techniques may be employed instead of the naive 
      stochastic gradient descent.
\item Stochastic action selection is not necessary 
      (except only at the early steps of each cycle). 
      In fact stochastic action selection may cause problems, 
      since it may lead to choices that are not suggested by the
      current policy \cite{Kon94}.
\item There is no need for a critic.
      The absence of a critic and the small number of  weight updates 
      contribute to the increase of the training speed.
\end{itemize}

The main disadvantage of the direct approach is that its performance
relies mainly on the effectiveness of the used optimization strategy. 
Due to the characteristics of the function to be optimized one cannot
be certain that any kind of optimization approach will be suitable for
training.
 
As already stated, a previous reinforcement learning
approach that follows a direct strategy, employs 
optimization techniques based on genetic algorithms and provides
very good results in terms of training speed (required number of cycles)
\cite{Whi93}. 
In this work, we present a different optimization strategy 
based on the polytope algorithm \cite{Neld65,Gil89,Nas96}, 
which is described next.

\subsection{The Polytope Algorithm}

The Polytope algorithm belongs to the class of direct 
search methods for non-linear optimization. It is also known by the name
Simplex, however it should not be confused with the well known Simplex method
of linear programming. Originally this algorithm was designed 
by Spendley et al.
\cite{Spe62} and was refined later by Nelder and Mead \cite{Neld65}.
A polytope (or simplex) in $R^{(n)}$ is a construct with $(n+1)$ vertices 
(points in $R^{(n)}$) defining
a volume element. For instance in two dimensions the simplex is a triange, in 
three dimensions it is a tetrahydron, and so on so forth. In our case
each vertex point $w_i=(w_{i1}, \ldots, w_{in})$ describes 
the $n$ parameters (weights and thresholds) of an action network. 

The input to the algorithm apart from a few parameters of minor importance,
is an initial simplex, ie $(n+1)$ points $w_i$. 
The algorithm brings the simplex in the area of a minimum,
adapts it to the local geometry, and finally shrinks it around the minimizer.
It is a derivative-free, iterative method and proceeds towards the minimum 
by manipulating a population of $n+1$ points 
(the simplex vertices) and hence it is expected
to be tolerant to noise, inspite its deterministic nature.
The steps taken in each iteration are described below. 
(We denote by $f$ the objective function and by $w_i$ the simplex vertices).

\begin{enumerate}
\item Examine the termination criteria to decide whether to stop or not.
\item Number the simplex vertices $w_i$, so that the sequence ${f_i=f(w_i)}$
      is sorted in ascending order.
\item Calculate the centroid of the first $n$ vertices: 
      $ c = \frac{1}{n}\Sigma_{i=0}^{n-1}w_i $
\item Invert the "worst" vertex $w_n$ as: 
      $ r = c +\alpha (c - w_n) $ (usually $\alpha =1$)
\item If $f_0 \le f(r) \le f_{n-1} $ then \\
      \hbox{\ \ \ }set $w_n = r, f_n= f(r)$, and go to step 1 \\
      endif
\item If $f(r) < f_0 $ then \\
      \hbox{\ \ \ }Expand as: $ e = c + \gamma (r-c)$ ($\gamma > 1$, usually $\gamma=2$) \\
      \hbox{\ \ \ }If $ f(e) < f(r) $ then \\
      \hbox{\ \ \ }\hbox{\ \ \ } set $w_n = e, f_n = f(e) $ \\
      \hbox{\ \ \ }else \\
      \hbox{\ \ \ }\hbox{\ \ \ }set $w_n = r, f_n = f(r) $ \\
     \hbox{\ \ \ }endif  \\
      \hbox{\ \ \ }go to step 1 \\
      endif
\item If $ f(r) \ge f_{n-1}$ then  \\
        \hbox{\ \ \ } If $ f(r) \ge f_n $ then \\
       \hbox{\ \ \ }\hbox{\ \ \ }contract as: $ k = c + \beta (w_n - c)$, 
        ($\beta < 1$, usually $\beta = \frac{1}{2}$) \\
        \hbox{\ \ \ } else \\
        \hbox{\ \ \ }\hbox{\ \ \ }contract as: $ k = c + \beta (r - c)$ \\
       \hbox{\ \ \ }  endif  \\
        \hbox{\ \ \ }If $ f(k) < min(f(r),f_n) $, then \\
        \hbox{\ \ \ }\hbox{\ \ \ }  set $w_n = k, f_n=f(k)$ \\
        \hbox{\ \ \ }else  \\
        \hbox{\ \ \ } \hbox{\ \ \ }Shrink the whole polytope as: \\
        \hbox{\ \ \ }\hbox{\ \ \ }Set $w_i = \frac{1}{2}(w_0+w_i),
         f_i = f(w_i)$ for $i = 1,2, \cdots , n$ \\
        \hbox{\ \ \ }endif \\
        \hbox{\ \ \ }go to step 1 \\
      endif
\end{enumerate}

In essence the polytope algorithm considers at each step a population
of $(n+1)$ action networks whose weight vectors $w_i$ are 
properly adjusted in order to obtain an action network with
high evaluation. In this sense, the polytope algorithm, 
although developed earlier, exhibits
an analogy with genetic algorithms which are also based
on the recombination of a population of points. 

The initial  simplex may be constructed in various ways.
The approach we followed was to pick the first vertex at random.
The rest of the vertices were obtained by line searches 
originating at the first vertex, along each of the $n$ directions.
This initialization scheme proved to be very effective for the pole balancing.
Other schemes such as, random initial vertices  or constrained  random vertices
on predefined directions, etc, did not work well. 
The termination criterion relies on comparing 
a measure for the polytope's "error" 
to a user preset small positive number. 
Specifically the algorithm returns if: \\
 $\frac{1}{n+1}\Sigma_{i=0}^n |f_i-\bar{f}| \le \epsilon $ 
where $\bar{f}=\frac{1}{n+1}\Sigma_{i=0}^n f_i $.

The use of the polytope algorithm has certain advantages like robustness
in the presence of noise, simple implementation and derivative-free operation.
These characteristics make this algorithm a suitable candidate
for use as an optimization tool in a direct reinforcement learning scheme.   
Moreover, since the method is deterministic, its effectiveness depends
partly on the initial weight values. For this reason, our training
strategy employs the polytope algorithm with random restarts 
as it will become clear in the application presented in the next section. 

It must also be stressed that the proposed technique does not make
any assumption concerning the architecture of the action network,
(which in the case described here is a multilayer perceptron),
and can be used  
with any kind of parameterized
action model (e.g. the fuzzy-neural action model employed in 
the GARIC architecture \cite{Ber92}).

\section{Application to the Pole Balancing Problem}

The pole balancing problem constitutes the best-studied 
reinforcement learning aplication. It consists of a single
pole hinged on a cart that may move left or right on a horizontal track
of finite length. The
pole has only one degree of freedom (rotation about the hinge point). The
control objective is to push the cart either left or right with a force
so that the pole remains balanced and the cart is kept
within the track limits. 

Four state variables are used to describe the status of the system at each
time instant: the horizontal position of the cart ($x$), the cart velocity
($\dot{x}$), the angle of the pole ($\theta$) and the angular velocity
($\dot{\theta}$). At each step the action network must decide the direction
and magnitude of force $F$ to be exerted to the cart. Details concerning
the equations of motion of the cart-pole system can be found in 
\cite{And89,Whi93,Lin96}.
Through Euler's approximation method we can simulate the cart-pole system
using discrete-time equations with time step $\Delta \tau = 0.02$ sec. 
We assume that the system's  equations of motion are not known 
to the controller, which perceives only the state vector at each time step.  
Moreover, we assume that a failure occurs when
$|\theta| > 12$ degrees or $|x| > 2.4m$ and that training has been
successfully completed if the pole remains balanced for more than
120000 consequtive time steps. 
Two versions of the problem exist concerning the magnitude of the
applied force $F$. We are concerned with the case where the
magnitude is fixed and the controller must
decide only the direction of the force at each time step. Obviously
the control problem is more difficult comprared to the case where 
any value for the magnitude is allowed. Therefore, comparisons will be
presented only with fixed magnitude approaches and we will not consider
architectures like the RFALCON \cite{Lin96}, which are very efficient
but assume continuous values for the force magnitude. 

The polytope method is embeded in the  MERLIN package \cite{Eva87,DGP}
for  multidimensional optimization.
Other derivative-free methods, provided by MERLIN 
have been tested in the pole-balancing example
(random, roll \cite{Eva87}), but the results
were not satisfactory. On the contrary, 
the polytope algorithm was very effective
being able to balance the pole in a relative few number 
of cycles (function evaluations) which was less than 1000 in many cases.
As mentioned in the previous section, the polytope method is 
deterministic, thus its effectiveness depends partly 
on the initial weight values. For this reason
we have employed an optimization strategy that is based on  
the polytope algorithm with random restarts. 
Each run starts by randomly specifying an initial point in the
weight space and constructing the initial polytope by performing
line minimizations along each of the $n$ directions. 
Next, the polytope algorithm  is run  for up to 100 function
evaluations (cycles) and the optimization progress is monitored. 
If a cycle has been found lasting more than 100 steps, application
of the polytope algorithm continues for additional 750 cycles, otherwise
we consider that the initial polytope was not proper and
a random restart takes place. A random restart is also performed 
when after the additional 750 function evaluations the algorithm
has not converged, i.e., a cycle has not been encountered lasting
for more than 120000 steps (this maximum value is suggested in 
\cite{And87,Whi93}).        
In the experiments presented in this article a maximum of 15 restarts
was allowed. The strategy was considered unsuccessfully terminated if
15 unsucessful restarts were performed or the total number of function
evaluations was greater than 15000.   

The above strategy was implemented using the MCL programming 
language \cite{Cha89} that is part of the MERLIN optimization environment.
The initial weight values at each restart 
were randomly selected in the range
$(-0.5, 0.5)$. Experiments were also conducted that considered the ranges
$(-1.0, 1.0)$ and $(-2.0, 2.0)$ and the obtained results were similar, showing
that the method exhibits robustness
as far as the initial weights are concerned.

\begin{table}
\begin{center}
\begin{tabular}{||r|c|c|c|c||} \cline{2-5}
     \multicolumn{1}{c|}{}
   & \multicolumn{4}{c||}{\em Number of Cycles} \\
     \hline
     \multicolumn{1}{||c|}{\em Method}
   & \multicolumn{1}{c|}{Best}
   & \multicolumn{1}{c|}{Worst}
   & \multicolumn{1}{c|}{Mean}
   & \multicolumn{1}{c||}{SD}  \\ \hline
   Polytope & 217  &  10453  & 2250  & 1955  \\ \hline
   AHC      & 4123  & 12895    & 6175  & 2284  \\ \hline
   GA-100   & 886  &  11481  & 4097  & 2205  \\ \hline
\end{tabular}
\end{center}
\caption{Training performance in terms of required number of training
cycles}
\end{table}

For comparison purposes the action network had also the same architecture 
with the architecture reported in \cite{Whi93,And89}. It is a multilayer
perceptron with four input units (accepting the sytem state),
one hidden layer with five sigmoid units and one sigmoid
unit in the output layer. There are also direct connections from the input units
to the output unit. The specification of the applied force characteristics from
the output value $y \in (0,1)$ was performed in the following way. At the
first ten steps of each cycle the specification was probabilistic, i.e.
$F=10N$ with probability equal to $y$. At the remaining steps the specification
was deterministic, i.e., if $y>0.5$ then $F=10N$, otherwise $F=-10N$. 
In this way, a degree of randomness is introduced in the function evaluation
process that assists in escaping from plateaus and shallow local minima.

Experiments have been conducted to assess the performance of the
proposed training method both in terms of training speed and generalization
capabilities. For comparison purposes we have also implemented the
AHC approach \cite{And87,And89},
while experimental results concerning the genetic reinforcement
approach on the same problem using the same motion equations and the
same network architecture are reported in \cite{Whi93}. Training speed
is measured in terms of the number of cycles (function evaluations) required
to achieve a successful cycle. A series of 50 experiments were conducted
using each method, with each cycle starting with random initial state variables.
Obtained results are summarized in Table 1, along with results 
from \cite{Whi93} concerning the genetic reinforcement case with population
of 100 networks (GA-100) that exhibited the best generalization performance. 
In accordance with previous published results, the AHC method does not manage
to find a solution in 14 of the 50 experiments (28\%), 
so the displayed results concern
values obtained considering only the successful experiments. 
On the contrary, the proposed training strategy was successful in all the
experiments and exhibited significantly better performance with respect to
the AHC case in terms of the required training cycles. 
From the displayed results it is also clear 
that the polytope method outperforms
the genetic approach, which is also better than the AHC method. 

\begin{table}
\begin{center}
\begin{tabular}{||r|c|c|c|c||} \cline{2-5}
     \multicolumn{1}{c|}{}
   & \multicolumn{4}{c||}{\em Percentage of Successful Tests} \\
     \hline
     \multicolumn{1}{||c|}{\em Method}
   & \multicolumn{1}{c|}{Best}
   & \multicolumn{1}{c|}{Worst}
   & \multicolumn{1}{c|}{Mean}
   & \multicolumn{1}{c||}{SD}  \\ \hline
   Polytope & 88.1  &  2.3  & 47.2  & 16.4  \\ \hline
   AHC      & 62.2  &  9.5  & 38.5  & 10.3  \\ \hline
   GA-100   & 71.4  &  3.9  & 47.5  & 14.2  \\ \hline
\end{tabular}
\end{center}
\caption{Generalization performance in terms of the percentage
of successful tests.}
\end{table}

Moreover, we have tested the generalization performance of the 
obtained action networks. These experiments are useful since a successful cycle
starting from an arbitrary initial position, does not nessecarily imply 
that the system will exhibit acceptable performance when started with 
different initial state vectors. The generalization experiments were
conducted following the guidelines suggested in \cite{Whi93}:
for each action network obtained in each 
of the 50 experiments
either using the polytope method or using the AHC method, a series of
5000 tests were performed from random initial states and we counted the
percentage of the tests in which the network was able to balance the pole
for more than 1000 time steps. 
The same failure criteria that were used for training
were also used for testing. Table 2 displays average results obtained by
testing the action networks obtained using the polytope 
and the AHC method (in the case of successful training experiments).
Moreover, it provides generalization results provided
in \cite{Whi93} concerning the GA-100 algorithm, using the same testing
criteria. As the results indicate the action networks obtained by all methods
exhibit comparative generalization performance. As noted in \cite{Whi93}
it is possible to increase the generalization performance by considering
stricter stopping criteria for the training algorithm. It must also be noted
that, in what concerns the polytope method, 
there was not any connection between
training time and generalization performance, i.e., the networks that
resulted by longer training times did not nessecarlily exhibit better
generalization capabilities.        

From the above results it is clear that direct approaches to delayed
reinforcement learning problems constitute a serious alternative to the
most-studied critic-based approaches. While critic-based approaches are
mainly based on their elegant formulation based on temporal differences and
stochastic dynamic programming, direct approaches base their success on
the power of the optimization schemes they employ. Such an optimization
scheme based on the polytope algorithm with random restarts 
has been presented in this work and was proved to be
very successful in dealing with the pole balancing problem. 
Future work will be focused 
on the employment of different 
kinds of action models (for example RBF networks) 
as well as the exploration of 
other derivative-free optimization schemes.    

One of us (I. E. L.) acknowledges 
partial support from the General 
Secretariat of Research  and  Technology under contract
PENED 91 ED 959.

\end{document}